\documentclass{article}

\usepackage{microtype}
\usepackage{graphicx}
\usepackage{subcaption}
\usepackage{booktabs} 
\usepackage{enumitem}

\usepackage{hyperref}

\usepackage[accepted]{icml2023}

\usepackage{amsmath}
\usepackage{amssymb}
\usepackage{mathtools}
\usepackage{amsthm}

\usepackage[capitalize,noabbrev]{cleveref}

\theoremstyle{plain}
\usepackage{shorthands}
\usepackage[textsize=tiny]{todonotes}

\icmltitlerunning{Text + Sketch: Image Compression at Ultra Low Rates}

\begin{document}

\twocolumn[
\icmltitle{Text + Sketch: Image Compression at Ultra Low Rates}



\icmlsetsymbol{equal}{*}

\begin{icmlauthorlist}
\icmlauthor{Eric Lei}{yyy}
\icmlauthor{Yi\u{g}it Berkay Uslu}{yyy}
\icmlauthor{Hamed Hassani}{yyy}
\icmlauthor{Shirin Saeedi Bidokhti}{yyy}

\end{icmlauthorlist}

\icmlaffiliation{yyy}{Department of Electrical and Systems Engineering, University of Pennsylvania, Philadelphia, PA, USA}

\icmlcorrespondingauthor{Eric Lei}{elei@seas.upenn.edu}

\icmlkeywords{Machine Learning, ICML}

\vskip 0.3in
]



\printAffiliationsAndNotice{} 

\begin{abstract}
    Recent advances in text-to-image generative models provide the ability to generate high-quality images from short text descriptions. These foundation models, when pre-trained on billion-scale datasets, are effective for various downstream tasks with little or no further training. A natural question to ask is how such models may be adapted for image compression. We investigate several techniques in which the pre-trained models can be directly used to implement compression schemes targeting novel low rate regimes. We show how text descriptions can be used in conjunction with side information to generate high-fidelity reconstructions that preserve both semantics and spatial structure of the original. We demonstrate that at very low bit-rates, our method can significantly improve upon learned compressors in terms of perceptual and semantic fidelity, despite no end-to-end training.  
\end{abstract}

\section{Introduction}
\label{sec:intro}

\begin{figure}[t]
     \centering
     \begin{subfigure}[b]{0.49\linewidth}
         \centering
        \includegraphics[width=\linewidth]{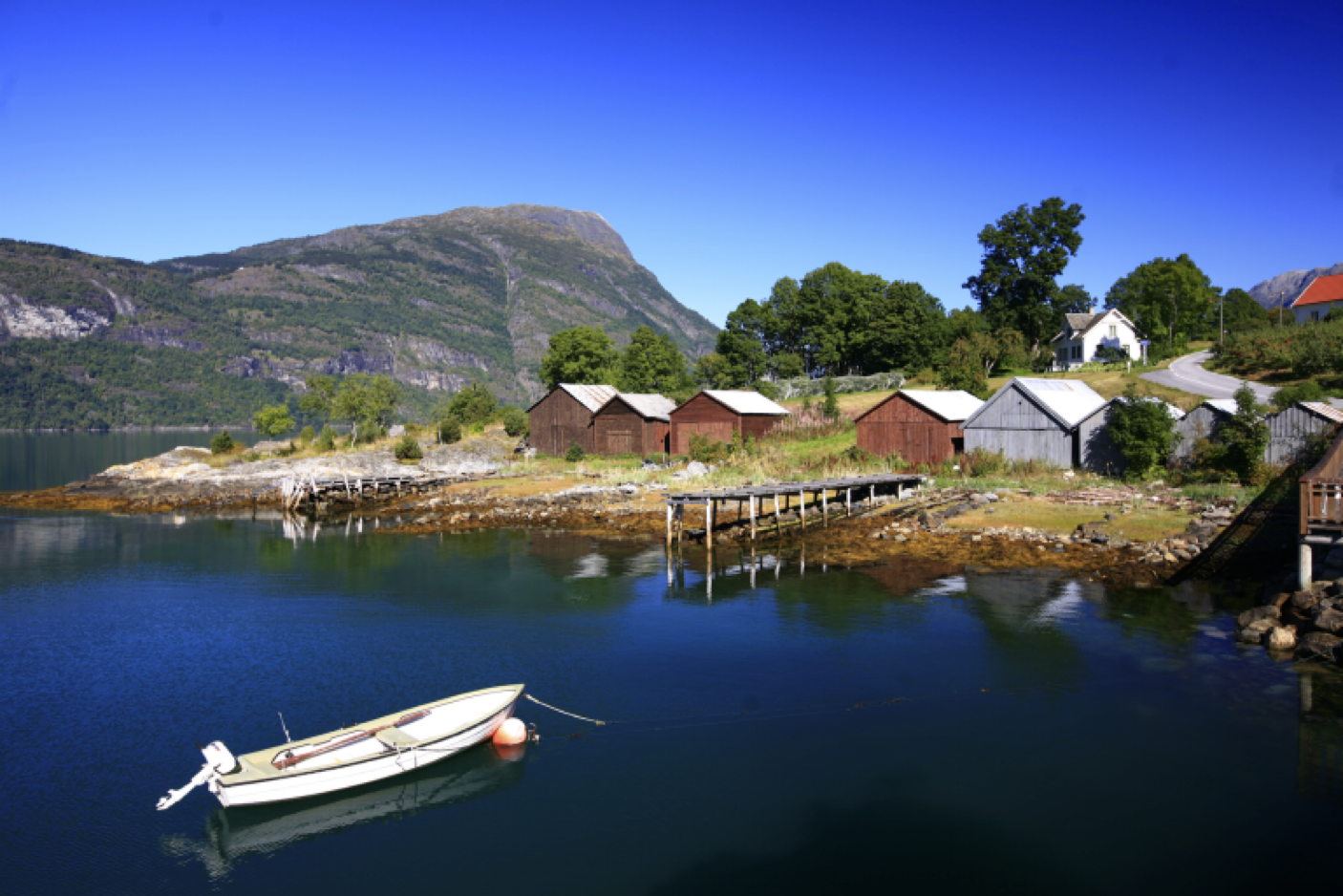}
        \caption{\scriptsize Ground-truth.}
     \end{subfigure}
     \vfill
     \begin{subfigure}[b]{0.49\linewidth}
         \centering
         \includegraphics[width=\linewidth]{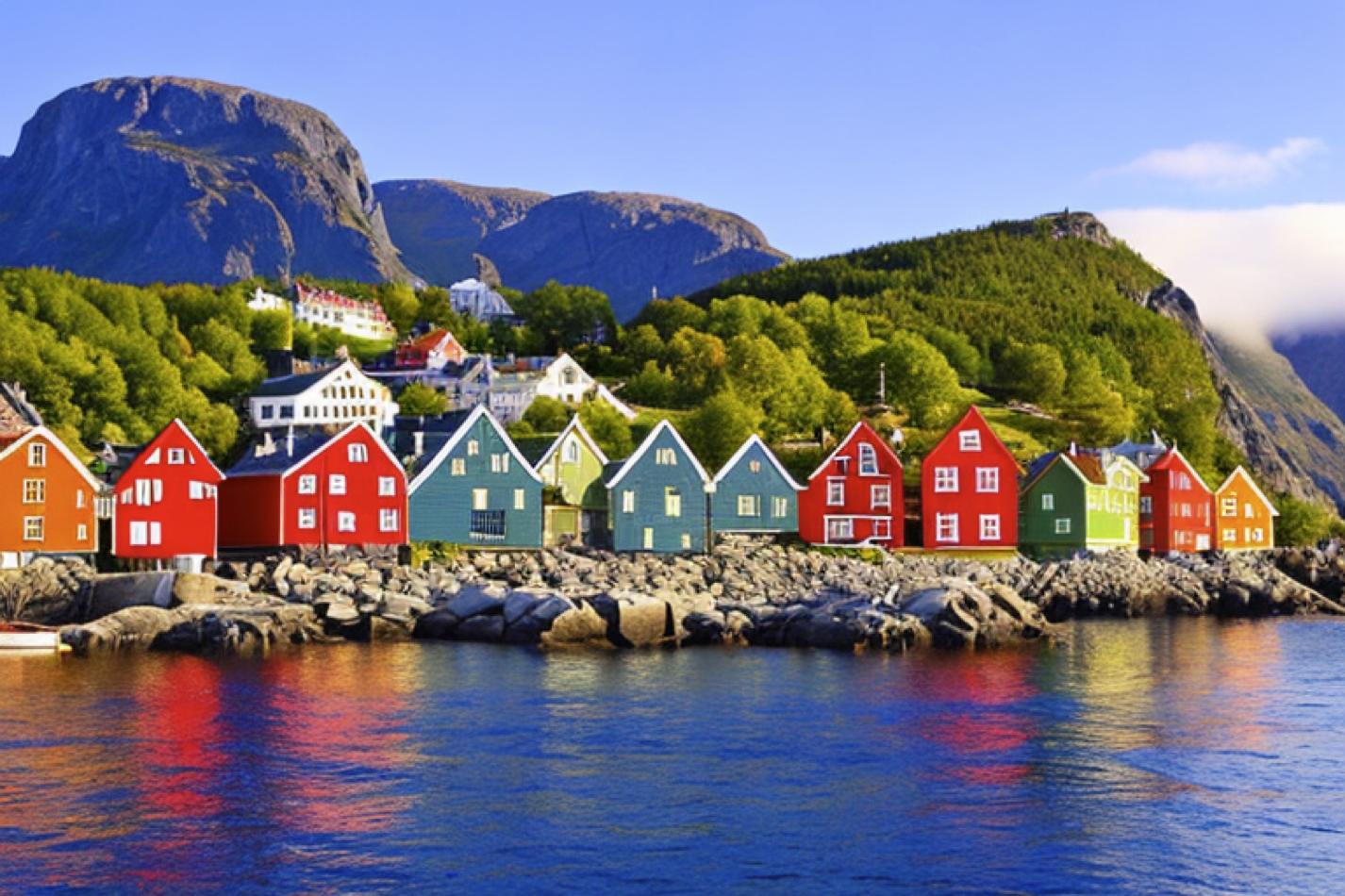}
         \caption{\scriptsize Text only reconstr. (0.0023 bpp).}
     \end{subfigure}
     \begin{subfigure}[b]{0.49\linewidth}
         \centering
         \includegraphics[width=\linewidth]{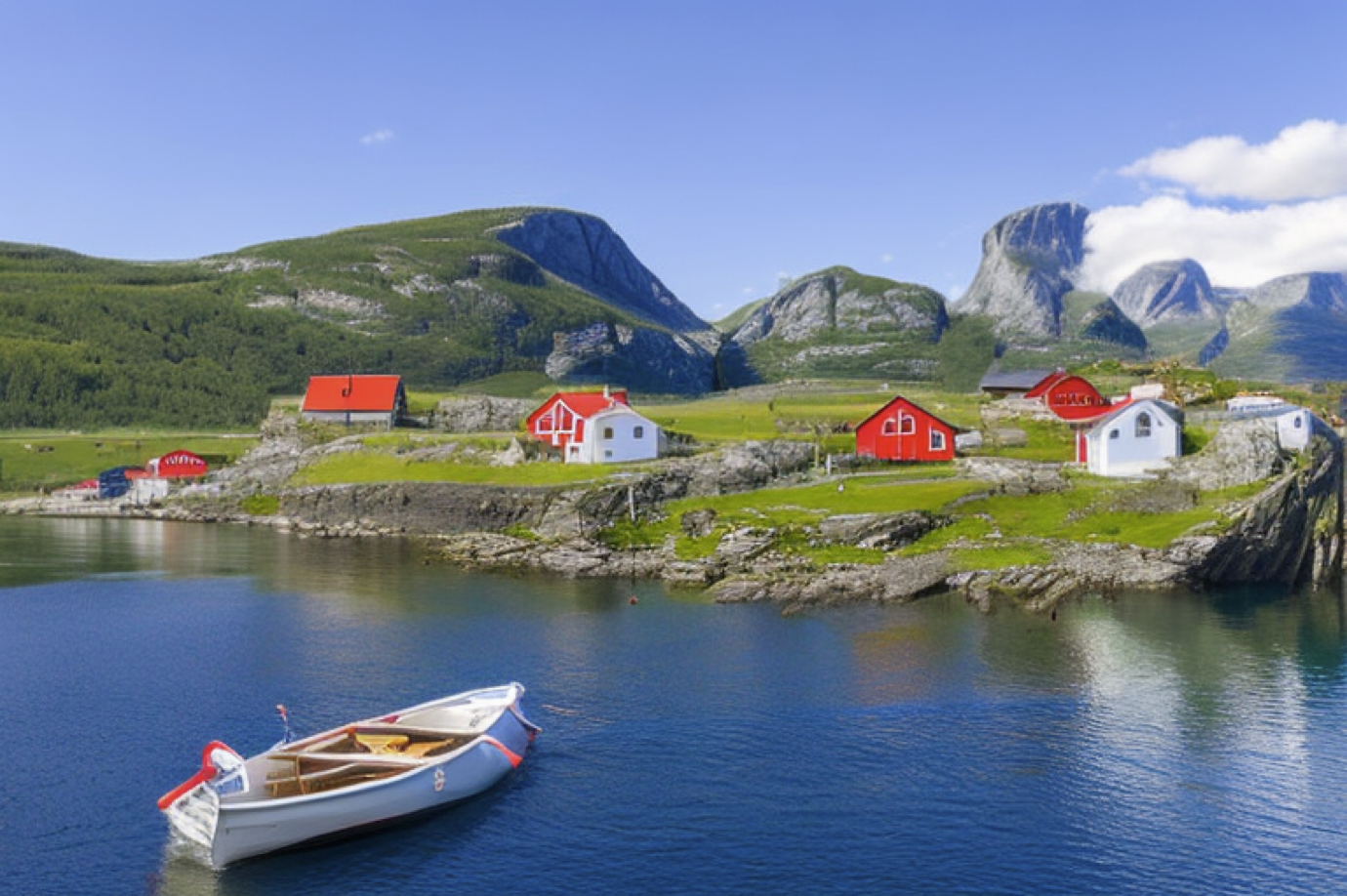}
         \caption{\scriptsize Text + sketch reconstr. (0.013 bpp).}
     \end{subfigure}
     \caption{Text-only reconstruction (PIC) preserves semantic information. Adding a sketch (PICS) preserves structural components.}
     \label{fig:CLIC7}
\end{figure}

Recent works from the lossy compression literature have demonstrated that when human satisfaction or semantic visual information is prioritized, compression schemes that manually encode images using human-written text descriptions as the compressed representation \cite{bhown2018towards, bhown2019dcc} yield significant improvements compared to traditional compressors. These works show that when operating at such low bit-rates, high levels of human satisfaction can still be achieved despite low pixel-wise fidelity. \cite{weissman2023toward} argues that transmitting the compressed information directly in the form of human language, known as textual transform coding, encodes information that scales with the semantic content in the image as interpreted by a human, rather than pixel-wise content.

Concurrent work in text-to-image generative models have provided the ability to generate high-quality images that represent the semantic information of the text across many domains \cite{ramesh2022hierarchical, rombach2022high}. These models, when scaled to orders of magnitude larger parameter counts and billion-scale datasets, have achieved remarkable capabilities in terms of converting language concepts to high quality images when assessed by humans. At such scale, these foundation models provide impressive zero-shot capabilities, allowing them to be used as a backbone when designing models for tasks not explicitly trained for. 

Prior neural compression paradigms, such as generative compression, attempt to align its reconstructions with human assessment at low bit-rates by enforcing a distribution matching constraint. In contrast, our work investigates neural compression schemes that target human satisfaction by directly transmitting text containing human-aligned semantic information. By leveraging the recent advances in pre-trained foundation models that operate with vision and language, we demonstrate how neural compression can benefit from the scale of such models, whereas similarly scaled neural compressors would require extensive resources to train end-to-end.

Directly using an off-the-shelf text-to-image model (with no further training) to implement a textual transform code can yield good results in terms of preserving coarse semantic information at very low bit-rates. However, current language-vision models, typically built on top of CLIP \cite{radford2021learning}, are limited in the amount of semantic concepts they can synthesize, especially pertaining to the spatial placement of objects. As shown in Fig.~\ref{fig:CLIC7}, when sending a text that is CLIP-optimized as the compressed representation, coarse semantic information is kept, but lower-level details of the image such as the placement of objects is poor. We show how transmitting limited side information in the form of a sketch can preserve lower-level structures. Our full contributions are as follows.

\begin{enumerate}[label=\arabic*)]
    \item We design a neural compressor that uses text-to-image models in a zero-shot manner to implement compression schemes preserving human semantics at rates below 0.003 bits-per-pixel (bpp), which is an order of magnitude lower than previously studied regimes.
    \item We show how side information in the form of a compressed spatial conditioning map can be used to provide the high-level structural information in the image along with a transmitted text caption, producing reconstructions that improve structural preservation.
    \item We show that our schemes outperform state-of-the-art generative compressors in terms of semantic and perceptual quality, despite no end-to-end training.
\end{enumerate}

\section{Related Work}
\paragraph{Neural Compression.}
The use of neural networks to design lossy compressors was initiated by merging quantization with autoencoder architectures \cite{toderici2016, balle2017endtoend, theis2017lossy, agustsson2017soft}. These models are traditionally trained with reference distortion metrics such as MSE, MS-SSIM \cite{wang2003}, and LPIPS \cite{zhang2018}. However, reconstructions suffer from blurriness at low bit-rates, motivating the field of generative compression \cite{agustsson2019generative, mentzer2020high}. In this field, distortion can be sacrificed for perceptual quality \cite{blau2019rethinking}, measured as alignment between source and reconstruction distributions. This improves human satisfaction in the rate regime of $<$0.1 bpp, compressors tuned for pixel-wise distortions fail to generate realistic reconstructions. At such low bitrates, pixel-wise fidelity metrics fail to align with human perception, since they largely focus on low-level details rather than the higher-level structures. Generative compression thus allows for realistic but not necessarily faithful (with regards to a distortion measure) reconstructions. However, it poses realism in terms of a distribution matching formulation which can offer some alignment with human satisfaction; textual transform coding attempts to directly encode the human-aligned semantic information in the form of language. 

\paragraph{Text-to-Image Models.}
 While many architectures have been studied for text-to-image generation, such as VAEs \cite{ramesh2021zero, ding2021cogview} and GANs \cite{gal2021stylegannada}, diffusion models have become the method of choice due to easier scaling to massive datasets \cite{rombach2022high, ramesh2022hierarchical}. These methods typically leverage CLIP \cite{radford2021learning}, a pre-trained model that provides a shared text-image embedding space, to retrieve an embedding corresponding to the input text. The diffusion model uses this embedding as a conditional input to denoise randomly sampled noise into an image corresponding to the text. Our work does not necessarily require diffusion models per se; it can use any foundation model that can generate images from text, pre-trained at scale. 

Diffusion-based neural compressors have also been investigated \cite{yang2022lossy, pan2022extreme}. Rather than transmit text, these models transmit a quantized embedding as the conditional input to the diffusion-based decoder. DiffC \cite{theis2022lossy} directly transmits pixels corrupted by noise in a diffusion process. Contrary to these models, our proposed compressor uses fully pre-trained text-to-image models, transmits text directly as a compressed representation for the conditional input, and utilizes a spatial conditioning input as side information.

\paragraph{Human Compression.}
\cite{bhown2018towards, bhown2019dcc} demonstrates a hand-crafted compression scheme in which humans write down text descriptions of the image to compress; the decoder consists of another human who has access to a database of images and image editing software. Human-rated scores for this scheme were higher than WebP at similar rates, despite the fact that the reconstructions may not necessarily be faithful at the pixel-level. Building off these results, \cite{weissman2023toward} conjectures that human satisfaction is a function of pixel-level fidelity with a semantic fidelity, which can be interpreted via human language. At large rates, pixel-wise fidelity dominates human satisfaction; at low rates, pixel-wise fidelity becomes less meaningful when compared to the ``textual'' information of the image.

\section{Transmitting Text With Side Information}
\begin{figure*}[t]
    \centering
    \includegraphics[width=0.9\linewidth]{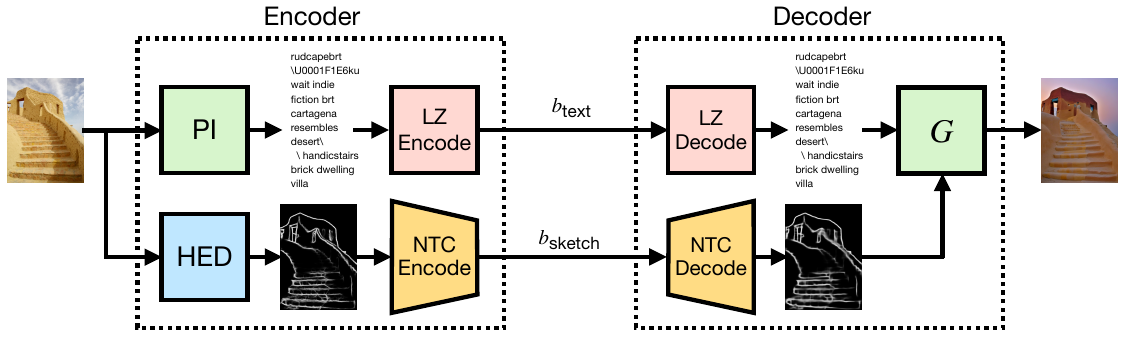}
    \caption{Diagram of PICS. Separate bitstreams for the text and sketch are losslessly encoded. Removing the bottom branch yields PIC. }
    \label{fig:PICS_diagram}
\end{figure*}

\subsection{Textual Transform Coding via Prompt Inversion}

Textual transform coding \cite{weissman2023toward} represents the image using a text description, which gets encoded with a lossless compressor. The decoder first recovers the text, which is used to synthesize the reconstructed image. Our decoder is assumed to be some text-to-image model $G$ that is pre-trained on a large-scale dataset. In this section, we use Stable Diffusion (SD) \cite{rombach2022high} for $G$.

One option to encode an image into text is via image captioning methods \cite{stefanini2022show}. However, most image captioning methods such as \cite{li2022blip} produce text captions that align with human language, but may not necessarily be optimal for the text-to-image model. Since SD uses pre-trained CLIP for text embeddings, it is more meaningful to directly search in the embedding space of CLIP in order to find text that represents an image for SD. 

Following \cite{wen2023hard}, we use prompt inversion (PI), which performs projected gradient search in CLIP's embedding space, using cosine similarity between the image embedding and the text embedding as the objective. To project to a hard text, the nearest CLIP embedding is found for each token being searched over. The tokens are converted to text and losslessly compressed. At the decoder, the decoded text is simply provided to $G$ which synthesizes a reconstructed image. We call this method Prompt Inversion Compression (PIC). PIC can achieve very low rates (around 0.002-0.003 bpp), yet preserve semantic information, since CLIP itself has semantic image comparison capabilities due to its vision-text merged feature space. 

An interesting fact of language-vision models such as SD is that quantization is naturally built into the model, where the language to vision conversion takes place. Text, after converted to tokens, is directly mapped to a codebook of embedding vectors. Thus, one can interpret prompt inversion as the encoder searching for the best CLIP codeword.

\subsection{Adding Spatial Conditioning Maps}
One challenge with using PIC is that it is difficult to increase reconstruction quality as the bitrate of text increases. As shown in \cite{wen2023hard}, increasing the number of tokens after a certain point fails to improve the CLIP score of the reconstructed image. Rather than attempting to increase the textual information in a way that $G$ can process, we instead propose to send side information in the form of a ``sketch'' of the original image, which contains finer structural information.

In this setting, we choose $G$ to be ControlNet \cite{zhang2023adding}, a text-to-image model built on top of SD that can process spatial conditioning maps in the form of edge detection maps, segementation maps, depth maps, etc. It ensures that the reconstructed images follow the spatial structure of the input map, and the style suggested by the text prompt. We use ControlNet as our decoder by sending a compressed version of the edge detection map (i.e., the sketch) as side information in addition to the prompt inversion text. In particular, we use the variant of ControlNet trained with Holistically-nested Edge Detection (HED) maps \cite{xie2015holistically} since those were found to have lower rate-distortion compared to Canny edge and segmentation maps. To compress the sketch, we use standard learned nonlinear transform codes (NTC) \cite{balle2020nonlinear} trained on a small dataset of HED maps. We call this scheme Prompt Inversion Compressor with Sketch (PICS), shown in Fig.~\ref{fig:PICS_diagram}.

\section{Experimental Results}

\begin{figure*}[!t]
     \centering
     \begin{subfigure}[b]{0.32\linewidth}
         \centering
        \includegraphics[width=0.8\linewidth]{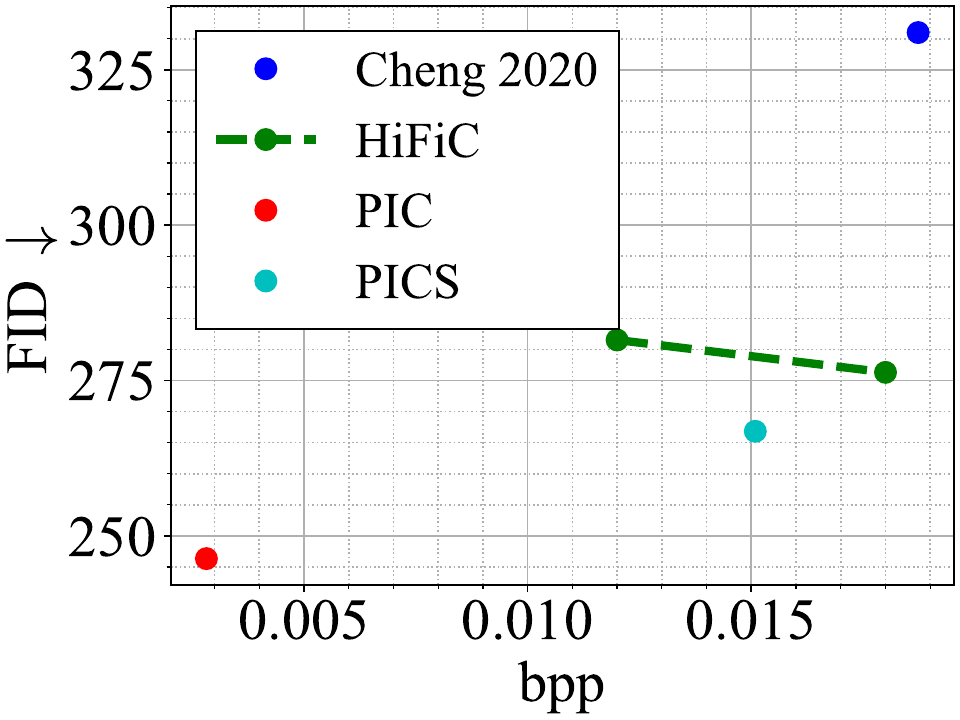}
     \end{subfigure}
     \begin{subfigure}[b]{0.32\linewidth}
         \centering
         \includegraphics[width=0.8\linewidth]{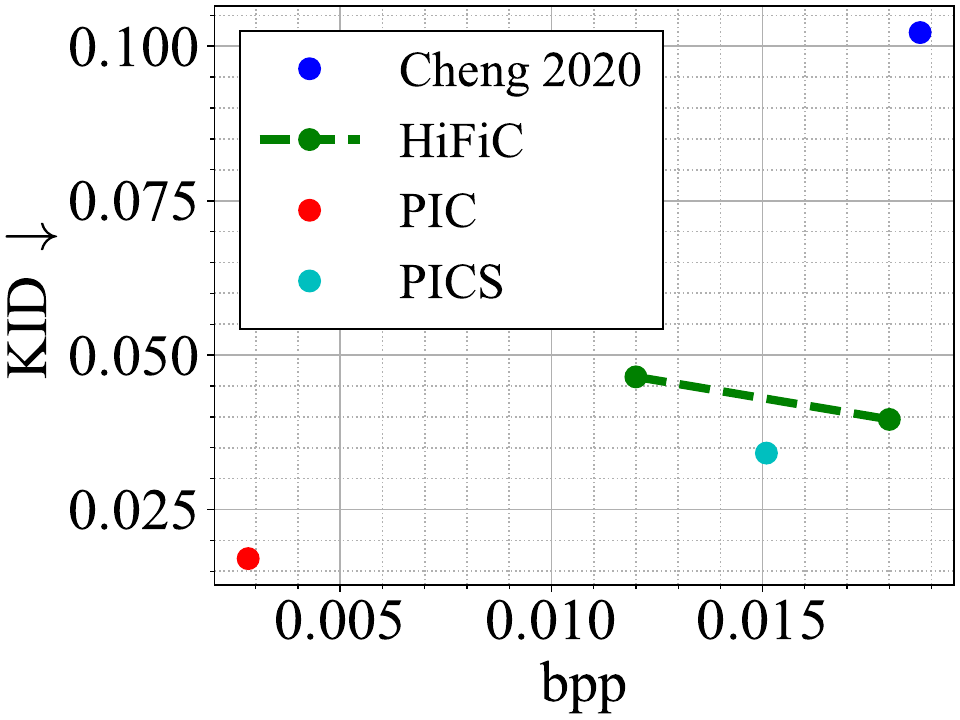}
     \end{subfigure}
     \begin{subfigure}[b]{0.32\linewidth}
         \centering
         \includegraphics[width=0.8\linewidth]{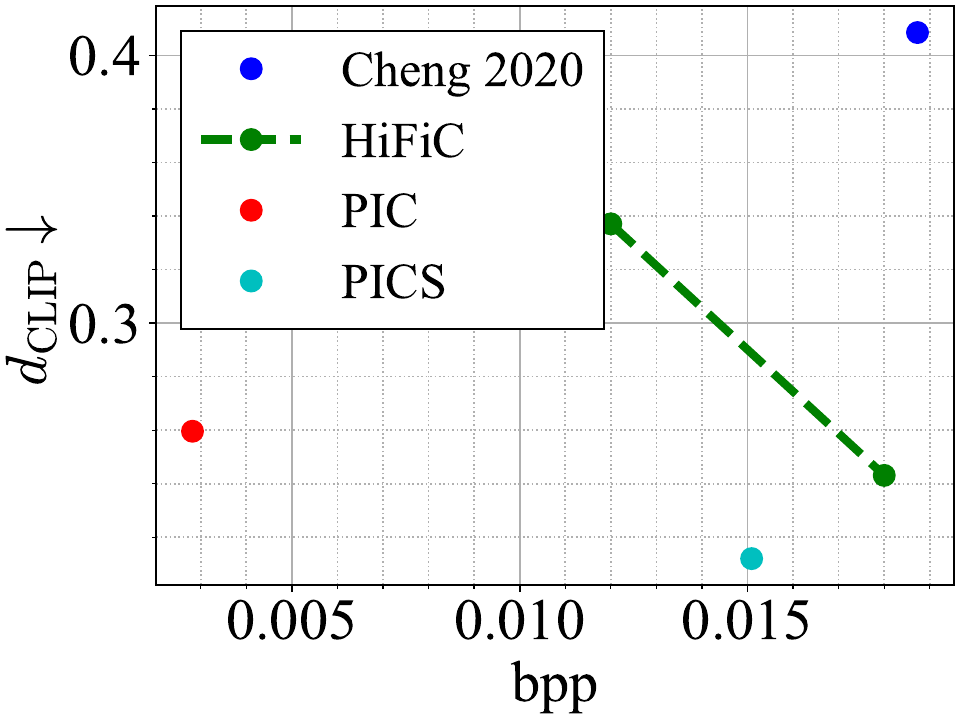}
     \end{subfigure}
     \caption{Achieved rate-perception and rate-distortion tradeoffs on CLIC 2021.}
     \label{fig:metricsCLIC}
\end{figure*}

\begin{figure*}[!t]
     \centering
     \begin{subfigure}[b]{0.32\linewidth}
         \centering
        \includegraphics[width=0.8\linewidth]{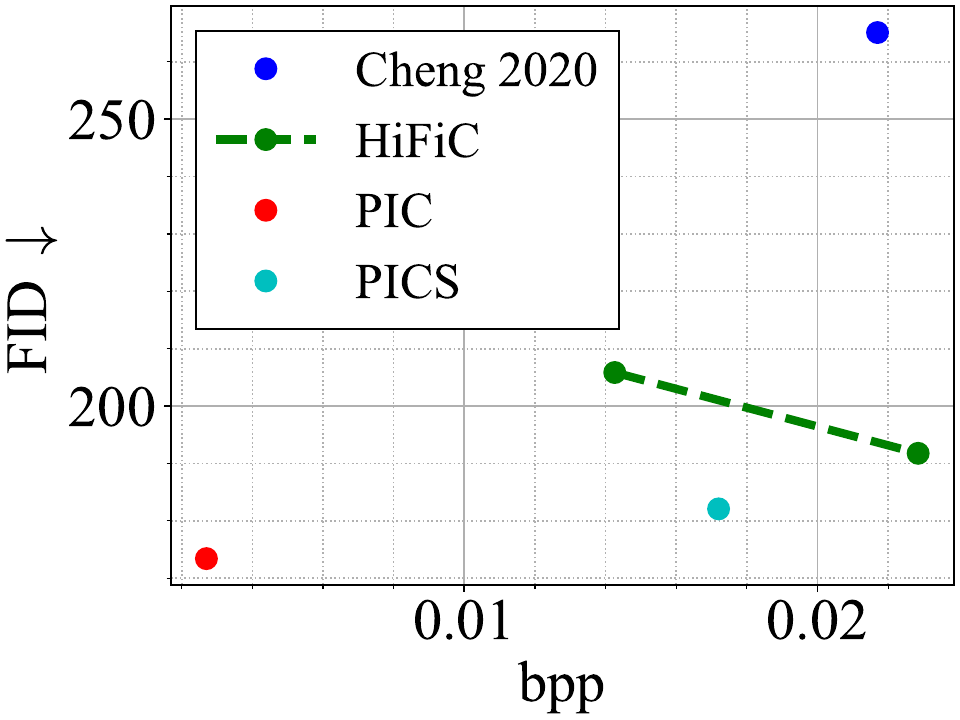}
     \end{subfigure}
     \begin{subfigure}[b]{0.32\linewidth}
         \centering
         \includegraphics[width=0.8\linewidth]{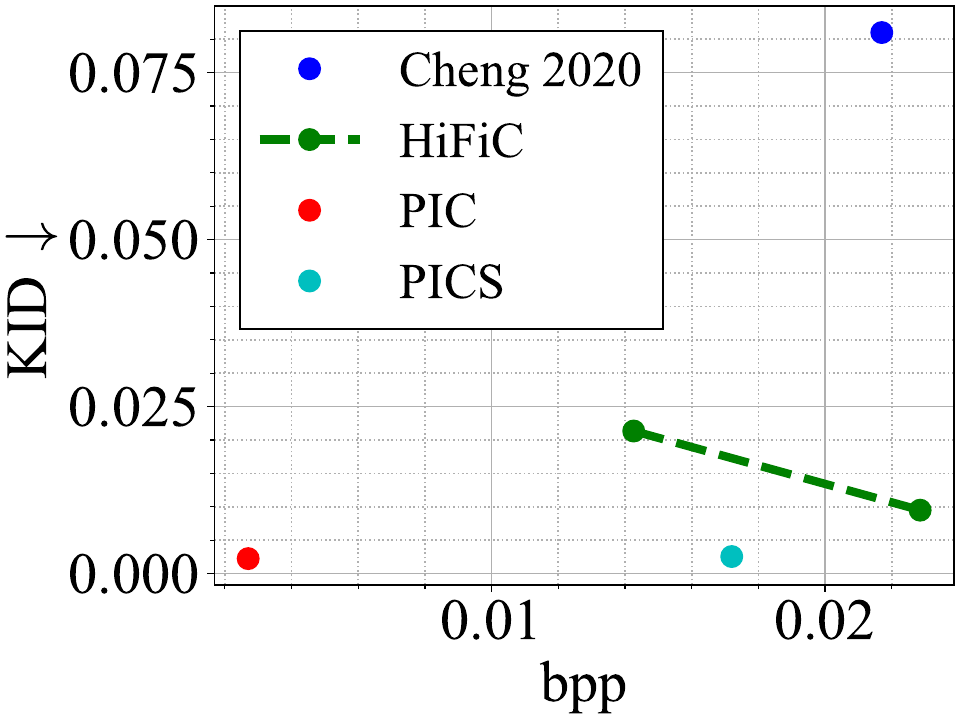}
     \end{subfigure}
     \begin{subfigure}[b]{0.32\linewidth}
         \centering
         \includegraphics[width=0.8\linewidth]{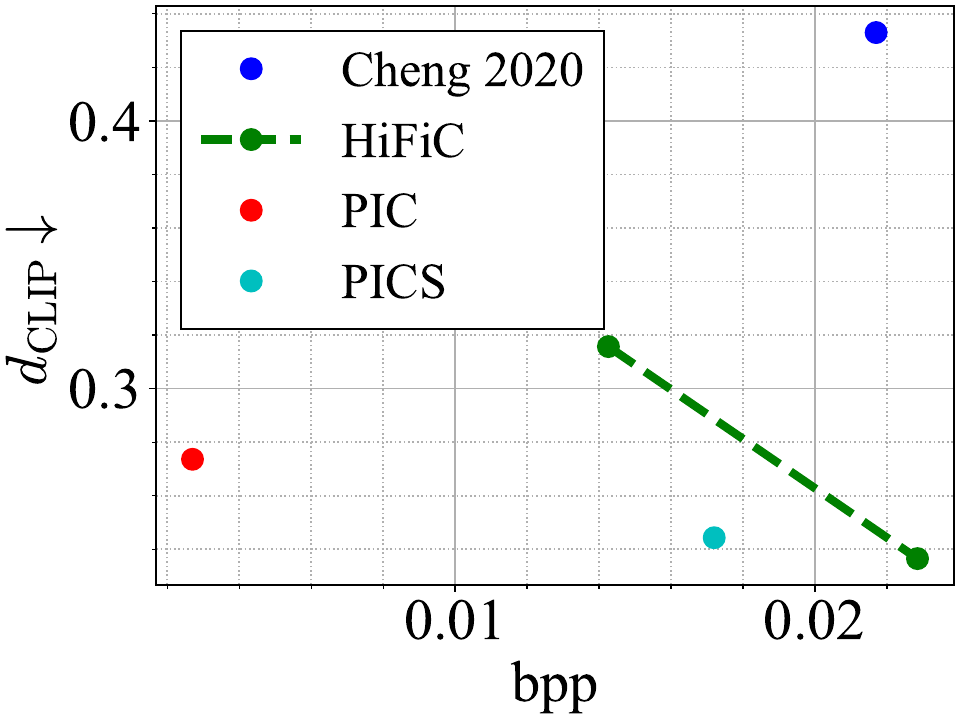}
     \end{subfigure}
     \caption{Achieved rate-perception and rate-distortion tradeoffs on DIV2K.}
     \label{fig:metricsDIV2K}
\end{figure*}

\begin{figure}[!t]
     \centering
     \begin{subfigure}[b]{0.49\linewidth}
         \centering
        \includegraphics[width=0.6\linewidth]{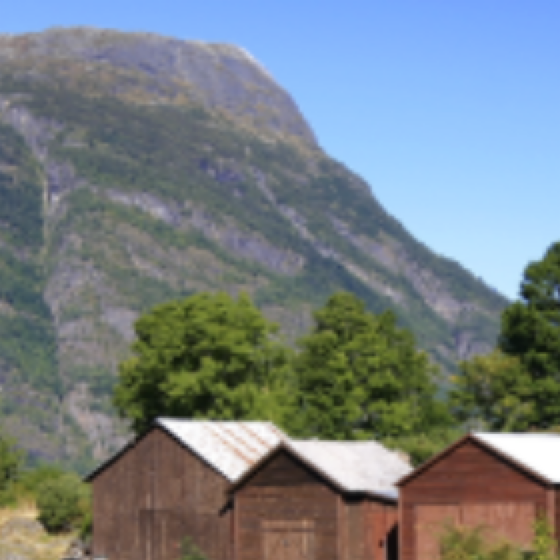}
        \caption{\scriptsize Ground-truth.}
     \end{subfigure}
     \begin{subfigure}[b]{0.49\linewidth}
         \centering
         \includegraphics[width=0.6\linewidth]{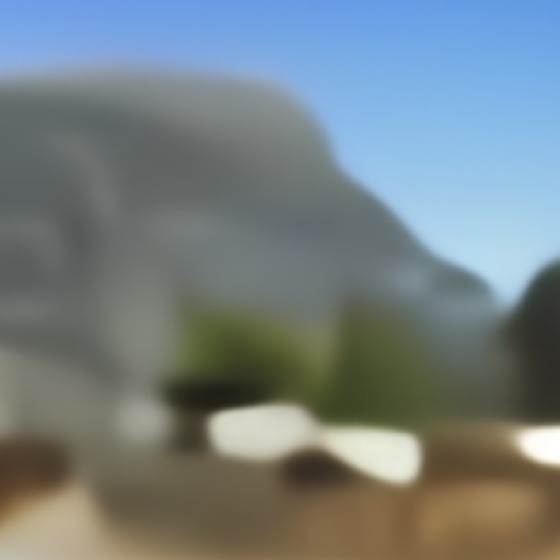}
         \caption{\scriptsize \cite{cheng2020learned} (0.018 bpp).}
     \end{subfigure}
     \begin{subfigure}[b]{0.49\linewidth}
         \centering
         \includegraphics[width=0.6\linewidth]{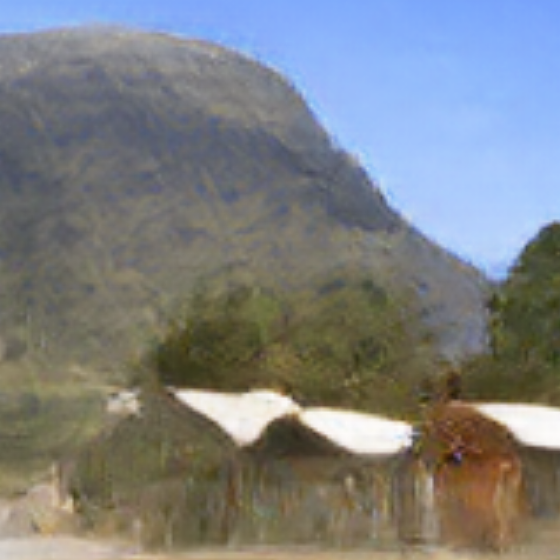}
         \caption{\scriptsize HiFiC (0.016 bpp).}
     \end{subfigure}
     \begin{subfigure}[b]{0.49\linewidth}
         \centering
         \includegraphics[width=0.6\linewidth]{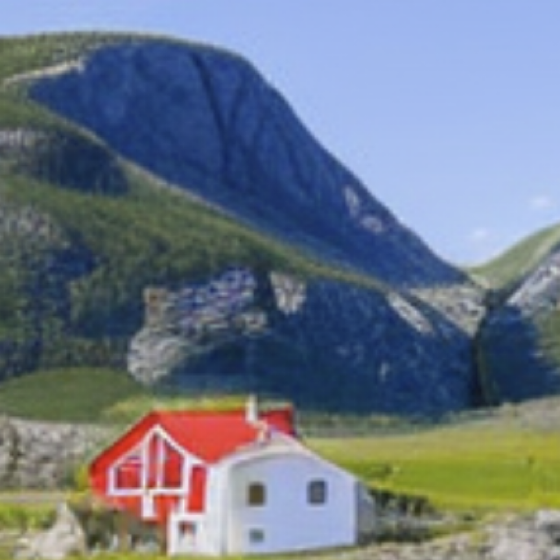}
         \caption{\scriptsize PICS (0.013 bpp).}
     \end{subfigure}
     \caption{Zoomed-in version of Fig.~\ref{fig:CLIC7}.}
     \label{fig:CLIC7zoom}
\end{figure}

\subsection{Setup}
\textbf{Datasets and Evaluation}: We use three evaluation datasets: Kodak \cite{Kodak}, CLIC 2021 \cite{CLIC} test, and DIV2K \cite{Agustsson2017DIV2K} validation. Since textual transform coding operates in an order of magnitude lower regime than even ``extreme'' compression ($<0.1$bpp), pixel-wise reference distortion metrics (PSNR, MS-SSIM, LPIPS) are not as meaningful. As human-aligned semantic reference metrics are still an open problem \cite{weissman2023toward}, we use cosine similarity of CLIP embeddings as a proxy,
\begin{equation}
    d_{\textsc{CLIP}}(\bx, \hat{\bx}) = 1 - \frac{e(\bx) \cdot e(\hat{\bx})}{\|e(\bx)\| \|e(\hat{\bx})\|},
\end{equation}
where $e(\cdot)$ is the image encoder of CLIP. Ideally, a human study would be performed, which we leave for future work. In addition, we use standard no-reference metrics to measure realism according to distributional alignment, FID \cite{heusel2017gans} and KID \cite{binkowski2018demystifying}. 

\textbf{Baseline Methods}: These include a generative compression baseline, HiFiC \cite{mentzer2020high}, and a NTC baseline \cite{cheng2020learned} optimized for MS-SSIM.

\textbf{PIC/PICS}: See appendix.

\subsection{Results}
\textbf{Quantitative Results}: Shown in Figs.~\ref{fig:metricsCLIC}, \ref{fig:metricsDIV2K}, we compare PIC/PICS in terms of rate-perception and distortion (in terms of $d_{\textsc{CLIP}}$). At such a low rate regime, HiFiC achieves better rate and semantic and perceptual quality than MS-SSIM trained NTC models. However, PICS is able to improve upon that further, with strict improvement in all tradeoffs. Interestingly, while PIC also strictly improves the rate-perception tradeoff, it performs worse in terms of semantic quality than PICS and HiFiC (albeit at lower rate). This shows that adding the sketch actually helps the generative model achieve higher semantic quality. 

\textbf{Qualitative Results}:
We visualize several reconstruction examples for all models and compare them with the ground-truths, in Figs.~\ref{fig:CLIC7}, \ref{fig:CLIC7zoom}, \ref{fig:CLIC56}, \ref{fig:kodim2}, and \ref{fig:Kodim19}. In general, PIC is able to reconstruct very coarse concepts contained in the ground-truth image. The NTC model optimized for rate-distortion yields blurry reconstructions in the low-rate regime. HiFiC improves realism, producing a sharper image with perhaps different textures than the original. In some cases, there are still compression artifacts, since HiFiC is not operating in the (near)-perfect realism regime. PICS is able to recover the high-level spatial structure of the ground-truth with superior sharpness, but synthesizes different textures or colors in the image. For example, Fig.~\ref{fig:CLIC7zoom} shows how PICS generates a house in front of a mountain of similar shape, but completely changes the color and style of the house as well as the composition of the mountainside. Additionally, PI-encoded prompts mostly recover semantic concepts, in Figs.~6-8.

\section{Conclusion}
In this paper, we use pretrained text-to-image models to construct a compressor that transmits a short text prompt and compressed image sketch. The only training required is to learn a lightweight learned compressor on HED sketches. Experimental results demonstrate superior performance in terms of semantic and perceptual quality. Current and future work includes a human study to evaluate human satisfaction of reconstructed images.
\bibliography{ref}
\bibliographystyle{icml2023}

\newpage
\appendix
\onecolumn
\section{Visual Reconstructions}
We place visual reconstructions referenced in the main text here. 
\begin{figure*}[t]
     \centering
     \begin{subfigure}[b]{0.325\linewidth}
         \centering
         \centering
        \includegraphics[width=0.85\linewidth]{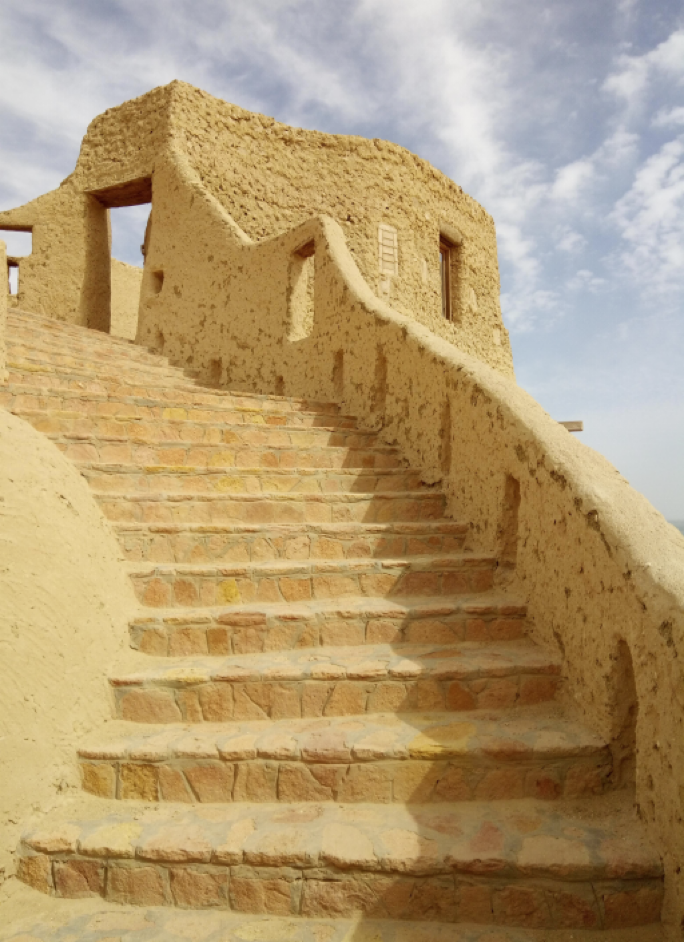}
        \caption{Ground-truth.}
     \end{subfigure}
     \begin{subfigure}[b]{0.325\linewidth}
         \centering
         \includegraphics[width=0.85\linewidth]{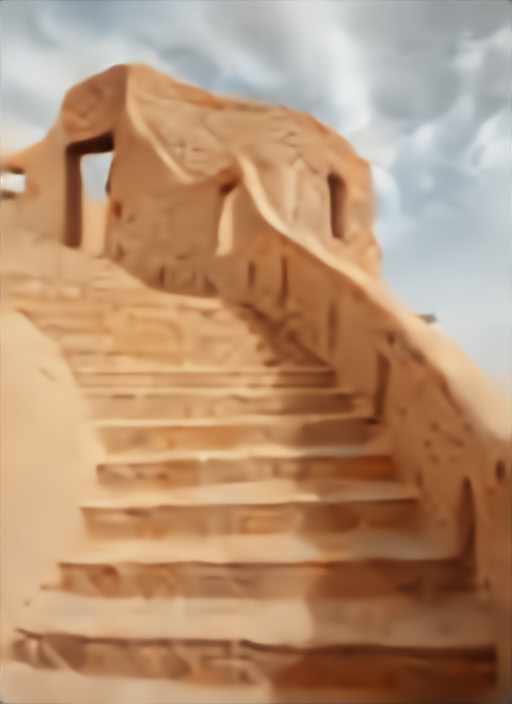}
         \caption{\cite{cheng2020learned} (0.015 bpp).}
     \end{subfigure}
     \begin{subfigure}[b]{0.325\linewidth}
         \centering
         \includegraphics[width=0.85\linewidth]{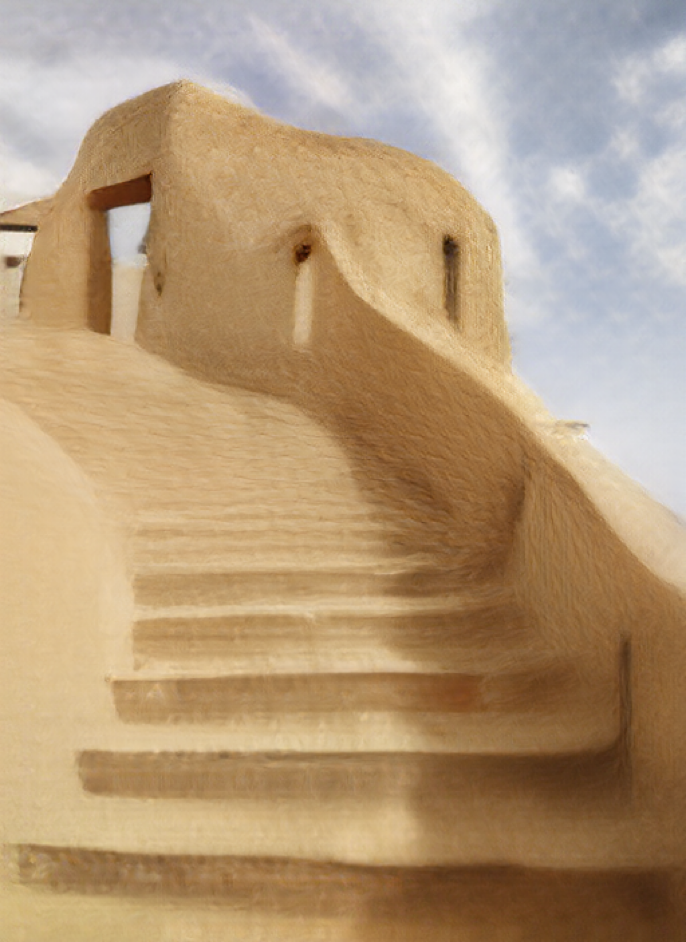}
         \caption{HiFiC reconstr. (0.013 bpp).}
     \end{subfigure}
     \begin{subfigure}[b]{0.325\linewidth}
         \centering
         \includegraphics[width=0.85\linewidth]{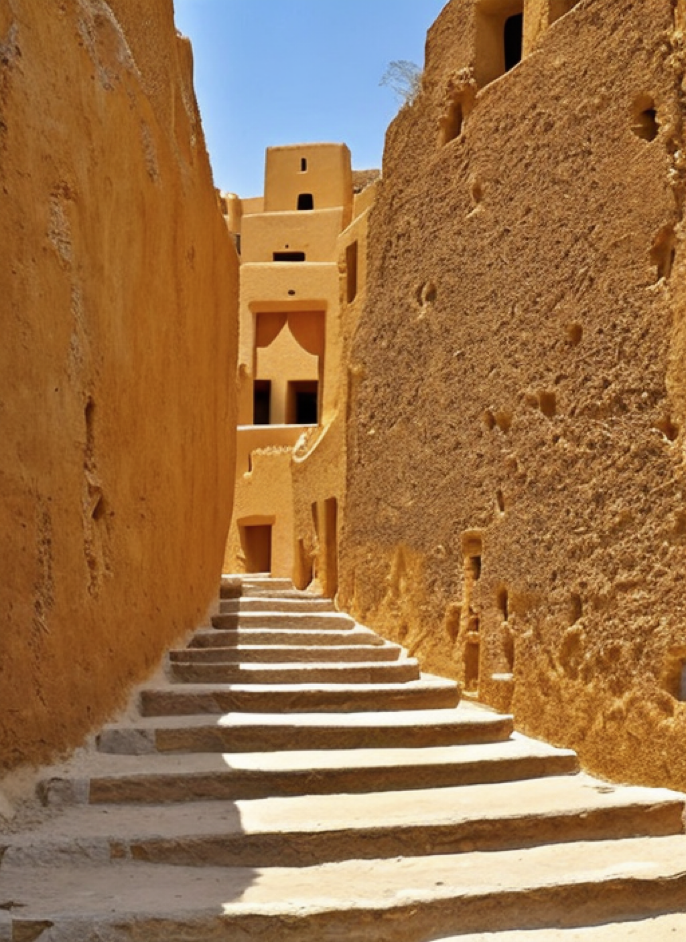}
         \caption{PIC reconstr. (0.0023 bpp).}
     \end{subfigure}
     \begin{subfigure}[b]{0.325\linewidth}
         \centering
         \includegraphics[width=0.85\linewidth]{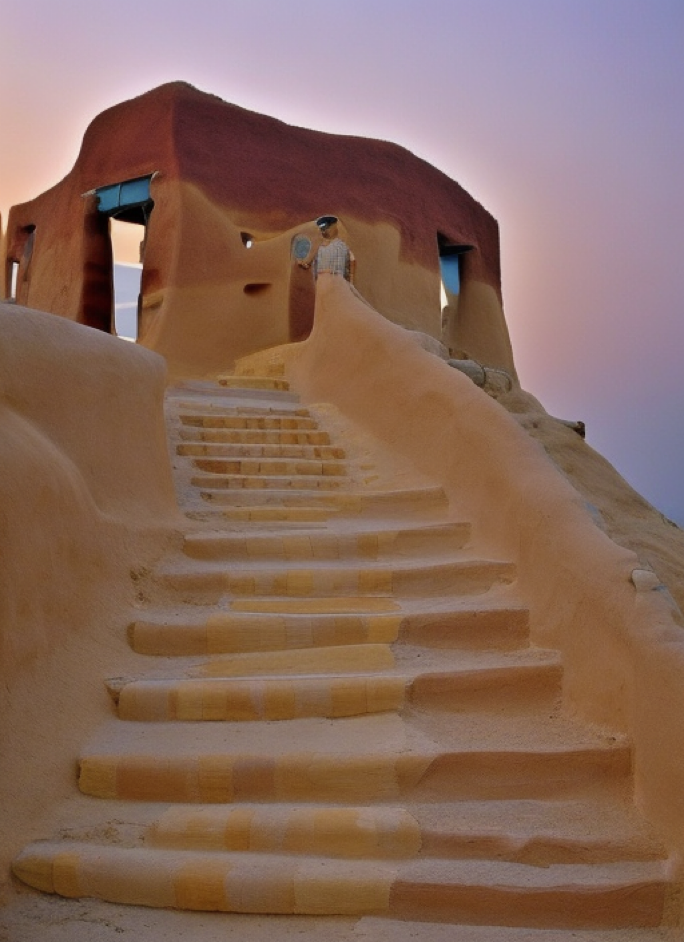}
         \caption{PICS reconstr. (0.013 bpp).}
     \end{subfigure}
     \begin{subfigure}[b]{0.325\linewidth}
         \centering
        \includegraphics[width=0.85\linewidth]{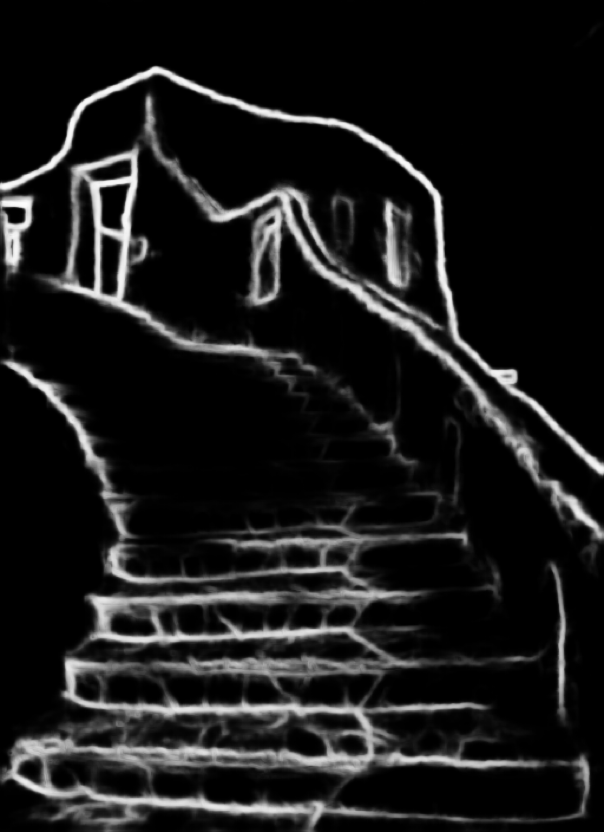}
        \caption{PICS sketch.}
     \end{subfigure}
     \begin{subfigure}[b]{0.325\linewidth}
         \centering
         \centering
        \includegraphics[width=0.85\linewidth]{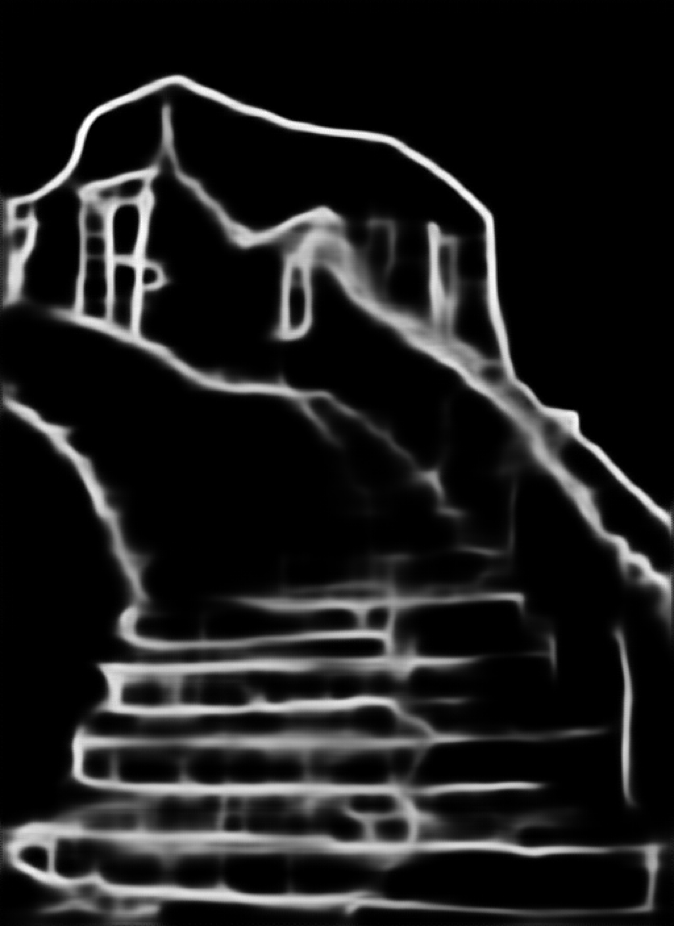}
        \caption{PICS sketch reconstr., 0.01 bpp.}
     \end{subfigure}
     \caption{From CLIC2021 test. For PIC/PICS, encoded prompt is ``rudcapebrt \symbol{92}U0001F1E6kuwait indie fiction brt cartagena resembles desert handicstairs brick dwelling villa''.}
     \label{fig:CLIC56}
\end{figure*}

\begin{figure*}[t]
     \centering
     \begin{subfigure}[b]{0.49\linewidth}
         \centering
         \centering
        \includegraphics[width=0.85\linewidth]{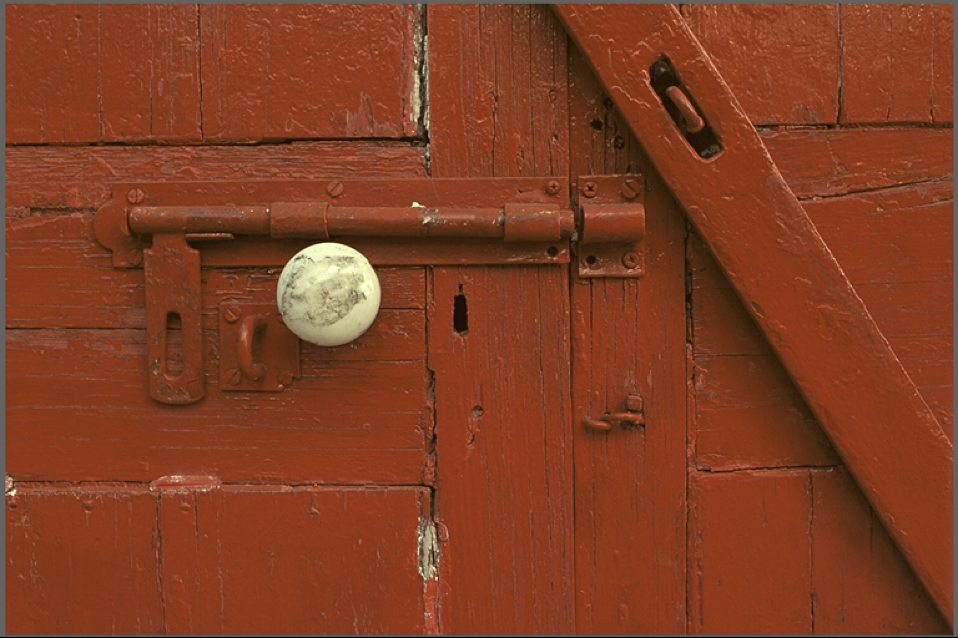}
        \caption{Ground-truth.}
     \end{subfigure}
     \begin{subfigure}[b]{0.49\linewidth}
         \centering
         \includegraphics[width=0.85\linewidth]{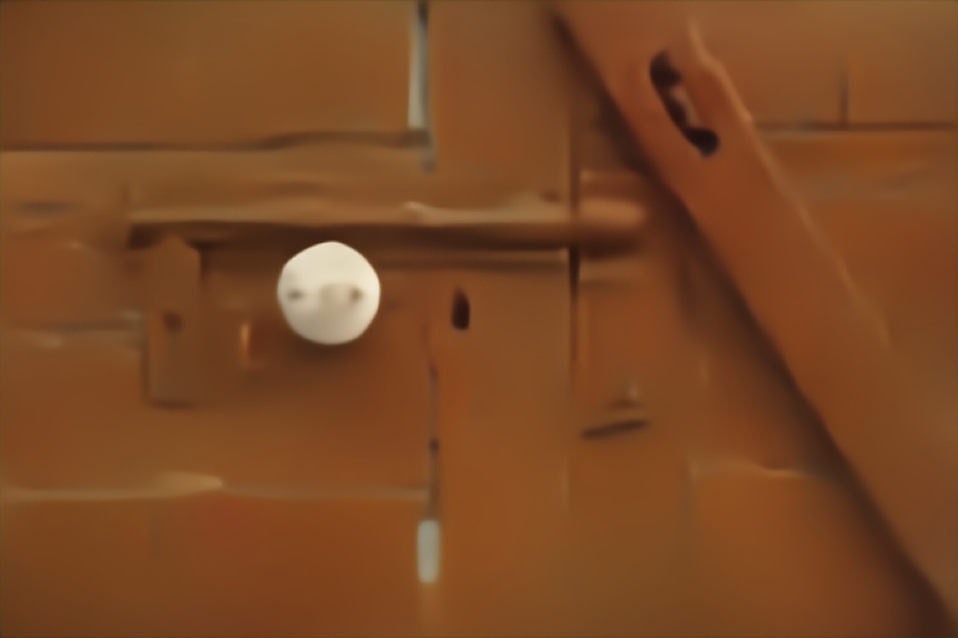}
         \caption{\cite{cheng2020learned} (0.013 bpp).}
     \end{subfigure}
     \begin{subfigure}[b]{0.49\linewidth}
         \centering
         \includegraphics[width=0.85\linewidth]{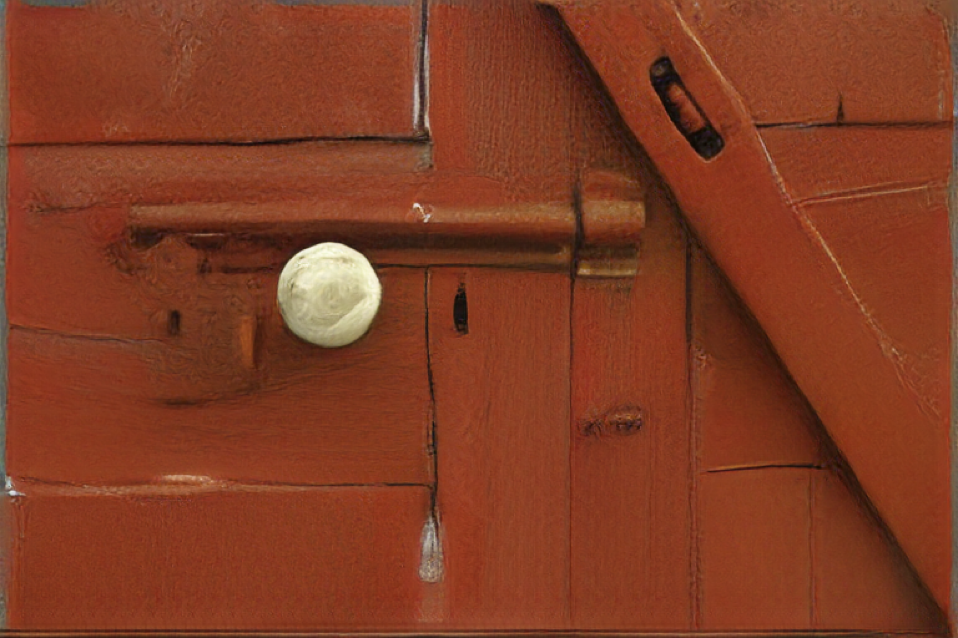}
         \caption{HiFiC reconstr. (0.020 bpp).}
     \end{subfigure}
     \begin{subfigure}[b]{0.49\linewidth}
         \centering
         \includegraphics[width=0.85\linewidth]{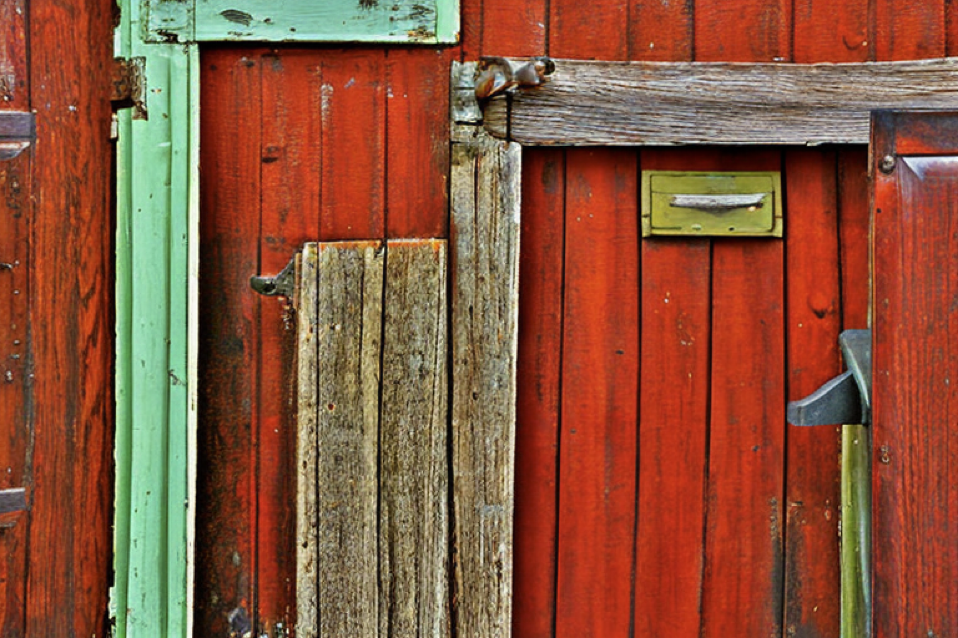}
         \caption{PIC reconstr. (0.0024 bpp).}
     \end{subfigure}
     \begin{subfigure}[b]{0.49\linewidth}
         \centering
         \includegraphics[width=0.85\linewidth]{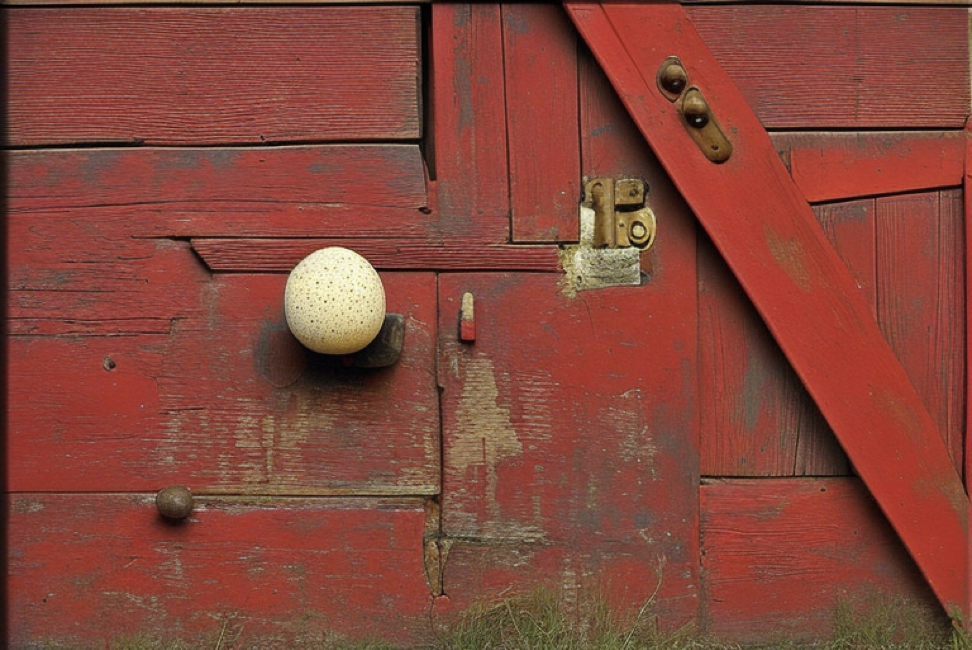}
         \caption{PICS reconstr. (0.011 bpp).}
     \end{subfigure}
     \begin{subfigure}[b]{0.49\linewidth}
         \centering
            \includegraphics[width=0.85\linewidth]{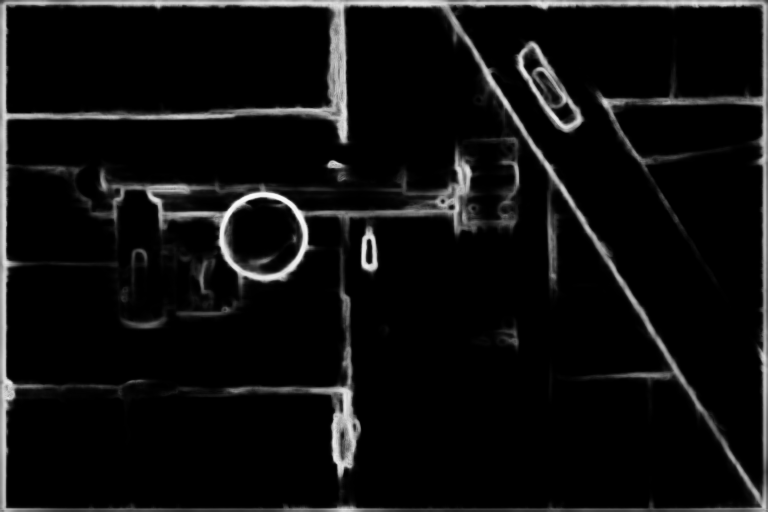}
        \caption{PICS sketch.}
     \end{subfigure}
     \begin{subfigure}[b]{0.49\linewidth}
         \centering
         \centering
        \includegraphics[width=0.85\linewidth]{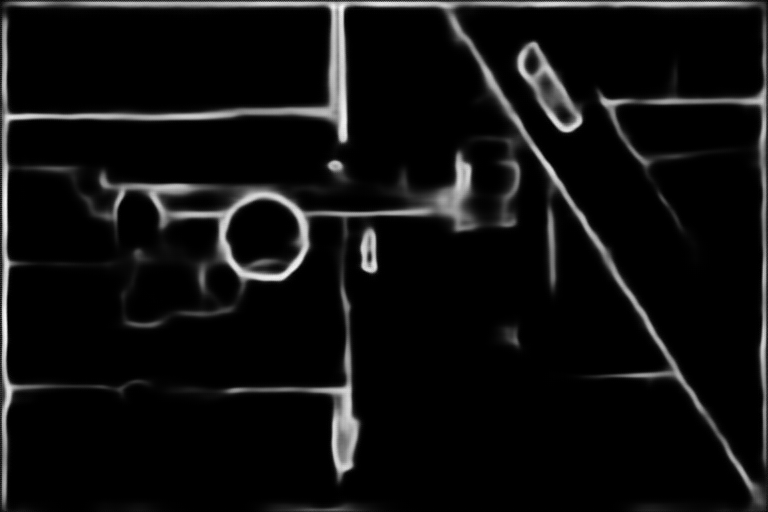}
        \caption{PICS sketch reconstr.}
     \end{subfigure}
     \caption{Kodim02. For PIC/PICS, encoded prompt is ``gayle chases eggs eggs knob withdrawn doors textures dewey red express u043D
  barns farcabs hauled''.}
     \label{fig:kodim2}
\end{figure*}

\begin{figure*}[t]
     \centering
     \begin{subfigure}[b]{0.325\linewidth}
         \centering
         \centering
        \includegraphics[width=0.8\linewidth]{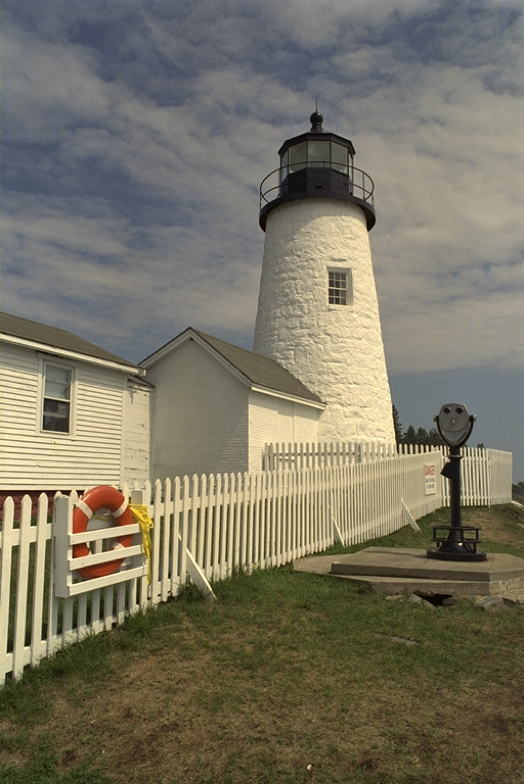}
        \caption{Ground-truth.}
     \end{subfigure}
     \begin{subfigure}[b]{0.325\linewidth}
         \centering
         \includegraphics[width=0.8\linewidth]{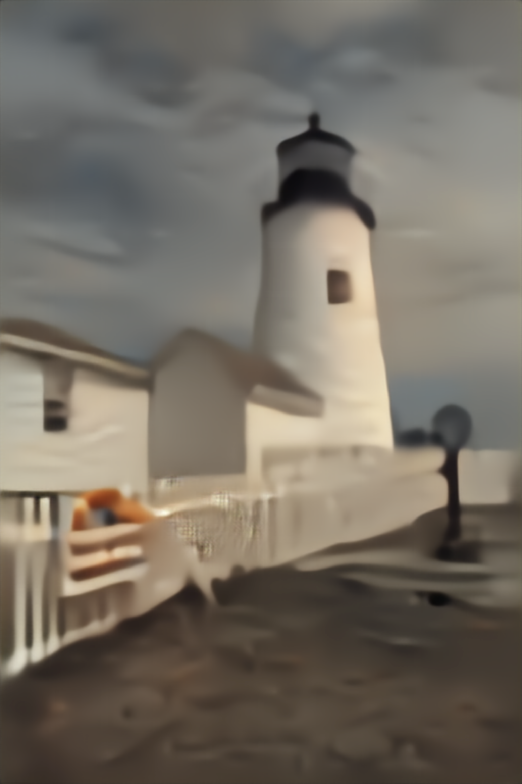}
         \caption{\cite{cheng2020learned} (0.016 bpp).}
     \end{subfigure}
     \begin{subfigure}[b]{0.325\linewidth}
         \centering
         \includegraphics[width=0.8\linewidth]{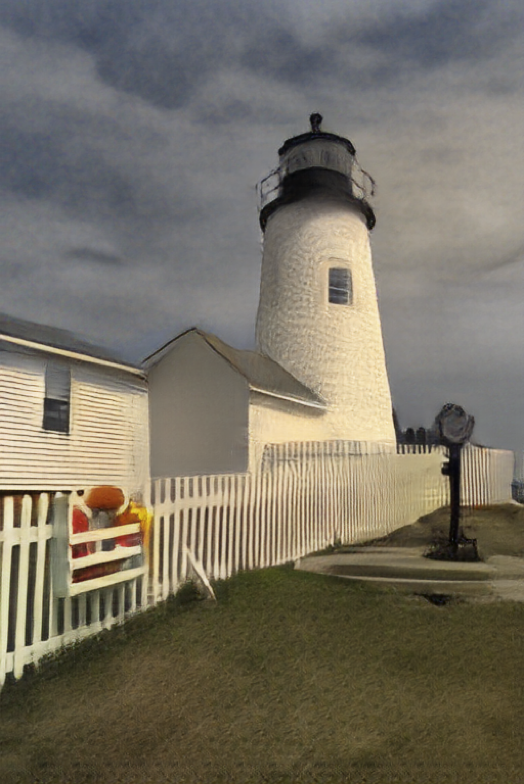}
         \caption{HiFiC reconstr. (0.028 bpp).}
     \end{subfigure}
     \begin{subfigure}[b]{0.325\linewidth}
         \centering
         \includegraphics[width=0.8\linewidth]{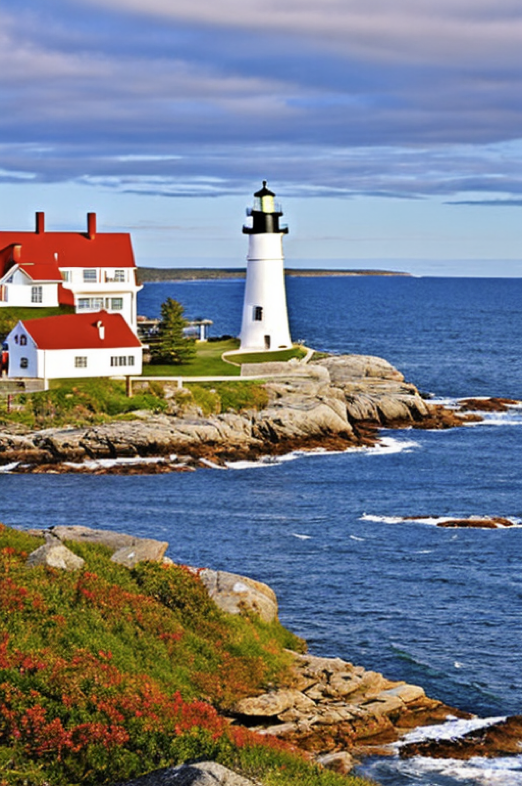}
         \caption{PIC reconstr. (0.0023 bpp).}
     \end{subfigure}
     \begin{subfigure}[b]{0.325\linewidth}
         \centering
         \includegraphics[width=0.8\linewidth]{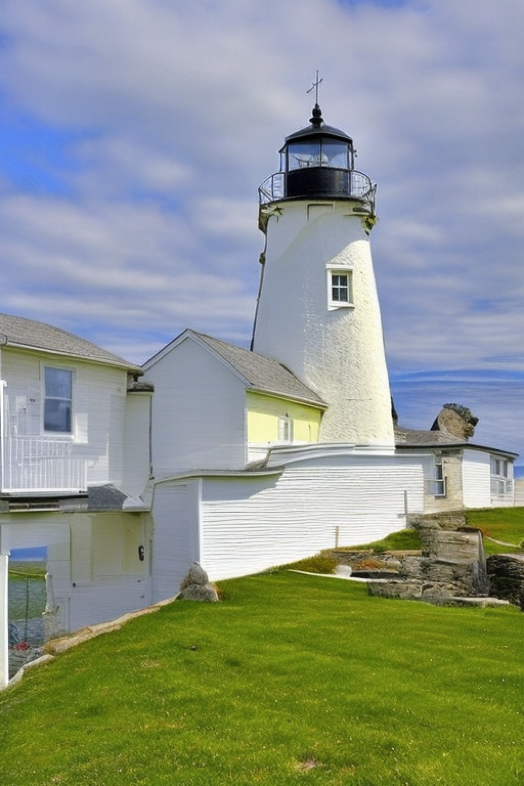}
         \caption{PICS reconstr. (0.012 bpp).}
     \end{subfigure}
     \begin{subfigure}[b]{0.325\linewidth}
         \centering
        \includegraphics[width=0.8\linewidth]{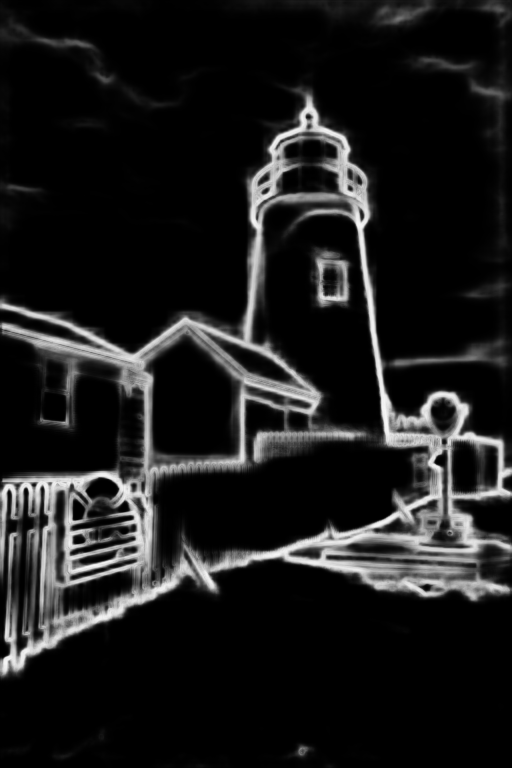}
        \caption{PICS sketch.}
     \end{subfigure}
     \begin{subfigure}[b]{0.325\linewidth}
         \centering
         \centering
        \includegraphics[width=0.75\linewidth]{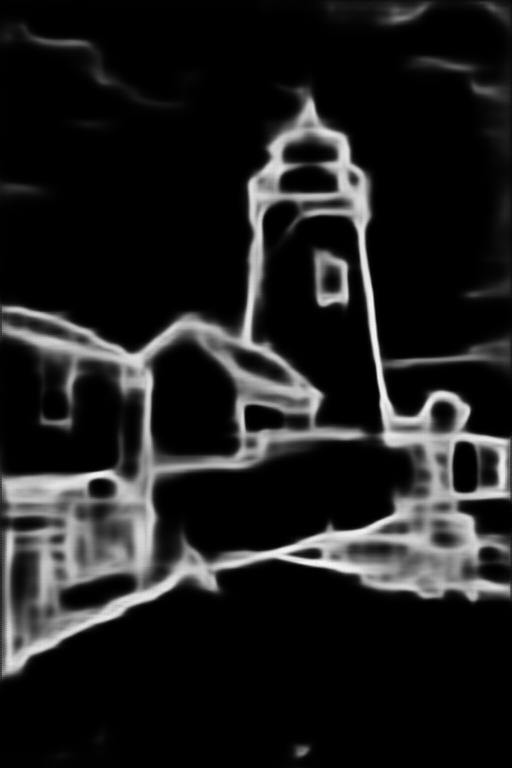}
        \caption{PICS sketch reconstr.}
     \end{subfigure}
     \caption{Kodim19. For PIC/PICS, encoded prompt is ``confederflipkquid rated confederfemale decorate maine k adm giggs ubunseeks
  lighthouse accomgigab''.}
     \label{fig:Kodim19}
\end{figure*}

\section{Implementation Details}
\subsection{Baselines and Evaluation}
 For HiFiC, we use an open-source implementation\footnote{\url{https://github.com/Justin-Tan/high-fidelity-generative-compression}} pre-trained on OpenImages \cite{kuznetsova2020open} to a target bitrate of 0.14 bpp. We then fine-tune on a subset of OpenImages with a target bitrate of 0.01 bpp, by setting $\lambda^{(a)} = 32, 64$. For the NTC baseline, we use a model pre-trained\footnote{\url{https://interdigitalinc.github.io/CompressAI/}} on Vimeo90K \cite{xue2019video}, fine-tuned on the same dataset for a target bitrate of 0.01 bpp. 

To compute FID and KID, we use the torch-fidelity\footnote{\url{https://github.com/toshas/torch-fidelity}} \cite{obukhov2020torchfidelity} package. 

\subsection{PIC/PICS}
For PI, we set the prompt length to 16 tokens, following the ablation study in \cite{wen2023hard}. To compress the HED sketch, we train a lightweight NTC model \cite{cheng2020learned} on HED maps from Vimeo90K under MS-SSIM distortion, targeting a bitrate of 0.01 bpp. We found that using MS-SSIM yielded better reconstructions from ControlNet compared to PSNR.

We use HuggingFace's diffusers library \cite{von-platen-etal-2022-diffusers} to run inference on SD and ControlNet. Although SD and ControlNet use many more parameters than NTC or HiFiC, one does not need to train these foundation models. Furthermore, with recent advances in efficient inference of diffusion models \cite{xFormers2022}, inference can be run efficiently on a single commodity GPU without using excessive memory. The code will be made available at \url{https://github.com/leieric/Text-Sketch}.

\end{document}